%% file: main.tex
\documentclass[10pt,conference]{IEEEtran}

\newif\iffinal
\finaltrue

\newif\ifieee
\ieeefalse

\newif\ifarxiv
\arxivtrue

\ifieee
    \ifarxiv
        \PackageError{Configuration}
        {IEEE and arXiv cannot both be enabled}
        {Set either \string\ieeefalse\space or \string\arxivfalse.}
    \fi
\fi

\usepackage{cite}
\usepackage{graphicx}
\DeclareGraphicsExtensions{.pdf,.png,.jpg,.jpeg}
\usepackage[cmex10]{amsmath}
\usepackage{amsthm}
\usepackage{array}
\usepackage[caption=false,farskip=0pt]{subfig}
\usepackage[hyphens]{url}
\usepackage[acronym]{glossaries}
\glsdisablehyper
\usepackage{xspace}
\usepackage{booktabs}
\usepackage[dvipsnames]{xcolor}
\usepackage[utf8]{inputenc}
\usepackage[T1]{fontenc}
\usepackage{multirow}
\usepackage{threeparttable}
\usepackage{balance}

\newcommand\major[1]{#1} %

\usepackage[colorlinks=true,allcolors=black]{hyperref} %

\usepackage[capitalise]{cleveref}

\newacronymstyle{long-short-br}
{%
  \GlsUseAcrEntryDispStyle{long-short}%
}%
{%
  \GlsUseAcrStyleDefs{long-short}%
}
\setacronymstyle{long-short-br}

\usepackage{transparent}
\usepackage{tikz}
\ifarxiv
    \newcommand\copyrighttext{%
      \scriptsize Accepted for presentation at SIBGRAPI 2026. The final published version will be available on IEEE~Xplore.}
    \newcommand\copyrightnotice{%
    \begin{tikzpicture}[remember picture,overlay]
    \node[anchor=south,yshift=30pt,xshift=0pt] at (current page.south) {\fbox{\transparent{0.85}\parbox{\dimexpr0.6\textwidth-\fboxsep-\fboxrule\relax}{\copyrighttext}}};
    \end{tikzpicture}%
    }
\else
\fi

\ifieee
\IEEEoverridecommandlockouts
\IEEEpubid{\makebox[\columnwidth]{979-8-3195-0255-1/26/\$31.00~\copyright2026 IEEE \hfill}
\hspace{\columnsep}\makebox[\columnwidth]{ }}
\else
\fi

\newcommand{\cmtid}{223}

\iffinal
\else
\usepackage[switch]{lineno}
\fi

\begin{document}

\title{Confidence-Aware Ensemble and Long-Word Refinement for Artistic Text Recognition}

\iffinal
    \author{
    \IEEEauthorblockN{Lucas A. Dias\IEEEauthorrefmark{1}, Henrique A. Schulz\IEEEauthorrefmark{1}, Rafaela de Miranda\IEEEauthorrefmark{1},\\Guilherme L. Peres\IEEEauthorrefmark{1}, Pedro L. Bittencourt\IEEEauthorrefmark{1}, and Rayson~Laroca\IEEEauthorrefmark{1}\\[0.75ex]}
    \IEEEauthorblockA{
        \IEEEauthorrefmark{1}\hspace{0.15mm}Pontifical Catholic University of Paran\'a, Curitiba, Brazil\\[0.75ex]
            \hspace{-0.75mm}\IEEEauthorrefmark{1}\hspace{-0.35mm}\tt{\small{\{dias.azevedo,henrique.schulz,r.miranda2,guilherme.peres,pedro.bittencourt\}}@pucpr.edu.br} \\ \IEEEauthorrefmark{1}{\tt\small rayson@ppgia.pucpr.br}}
    }
\else
    \author{SIBGRAPI Paper ID: \cmtid \\[7ex]}
    \linenumbers
\fi

\maketitle

\ifarxiv
    \copyrightnotice
\else
\fi

\input{0-acronyms}
\input{0-abstract}

\IEEEpeerreviewmaketitle

\input{1-intro}
\input{2-related}
\input{3-method}
\input{4-results}

\input{6-conclusions}
\input{0-acknowledgments}

\balance

\bibliographystyle{IEEEtran}
\bibliography{bibtex}

\end{document}

%% file: 0-acronyms.tex
\newacronym{atr}{ATR}{Artistic Text Recognition}
\newacronym{str}{STR}{Scene Text Recognition}
\newacronym{wra}{WRA}{Word Recognition Accuracy}
\newacronym{icdar}{ICDAR}{International Conference on Document Analysis and Recognition}
\newacronym{eccv}{ECCV}{European Conference on Computer Vision}
\newacronym{vit}{ViT}{Vision Transformer}
\newacronym{vlm}{VLM}{Vision-Language Model}
\newacronym{ctc}{CTC}{Connectionist Temporal Classification}
\newacronym{lm}{LM}{Language Model}
\newacronym{ar}{AR}{Autoregressive}
\newacronym{mae}{MAE}{Masked Autoencoder}
\newacronym{parseq}{PARSeq}{Permuted Autoregressive Sequence Model}
\newacronym{maerec}{MAERec}{Masked Autoencoder Recognition Model}
\newacronym{svtrv2}{SVTRv2}{Single Visual Text Recognition Version 2}
\newacronym{vitstr}{ViTSTR}{Vision Transformer for Scene Text Recognition}

%% file: 0-abstract.tex
\begin{abstract}
\gls*{atr} remains challenging because word images often combine decorative fonts, curved layouts, object-like characters, clutter, and severe distortions.
This paper studies WordArt-V1.5 as a standardized benchmark for this setting and evaluates recent scene and artistic text recognizers under a common protocol.
We propose a confidence-aware ensemble that combines SVTRv2, PARSeq, and MAERec after fine-tuning on the official training split.
The ensemble selects predictions using the minimum confidence over disagreement positions, emphasizing characters that separate competing hypotheses.
For long words, where a single character error can invalidate the whole prediction, we add a targeted refinement stage based on Needleman-Wunsch alignment and lexicon-guided correction.
On the WordArt-V1.5 Test~B split, the proposed system reaches 89.90\% Word Recognition Accuracy, improving the best individual fine-tuned model by 1.77 percentage points.
The long-word refinement produces a modest global gain, but improves the targeted long-word subset by 2.72 percentage points.
Finally, an error analysis of all remaining mistakes shows that 48.8\% are associated with labeling issues, visual ambiguity, or illegible samples, highlighting the value of diagnostic reporting for future ATR benchmarks and models.
Our source code is available at \textit{\url{https://github.com/lucas-azdias/Artistic-Text-Recognition/}}.
\end{abstract}

%% file: 1-intro.tex
\section{Introduction}
\label{sec:introduction}

\glsresetall

\begin{figure}[!t]
    \centering
    \includegraphics[width=0.80\linewidth]{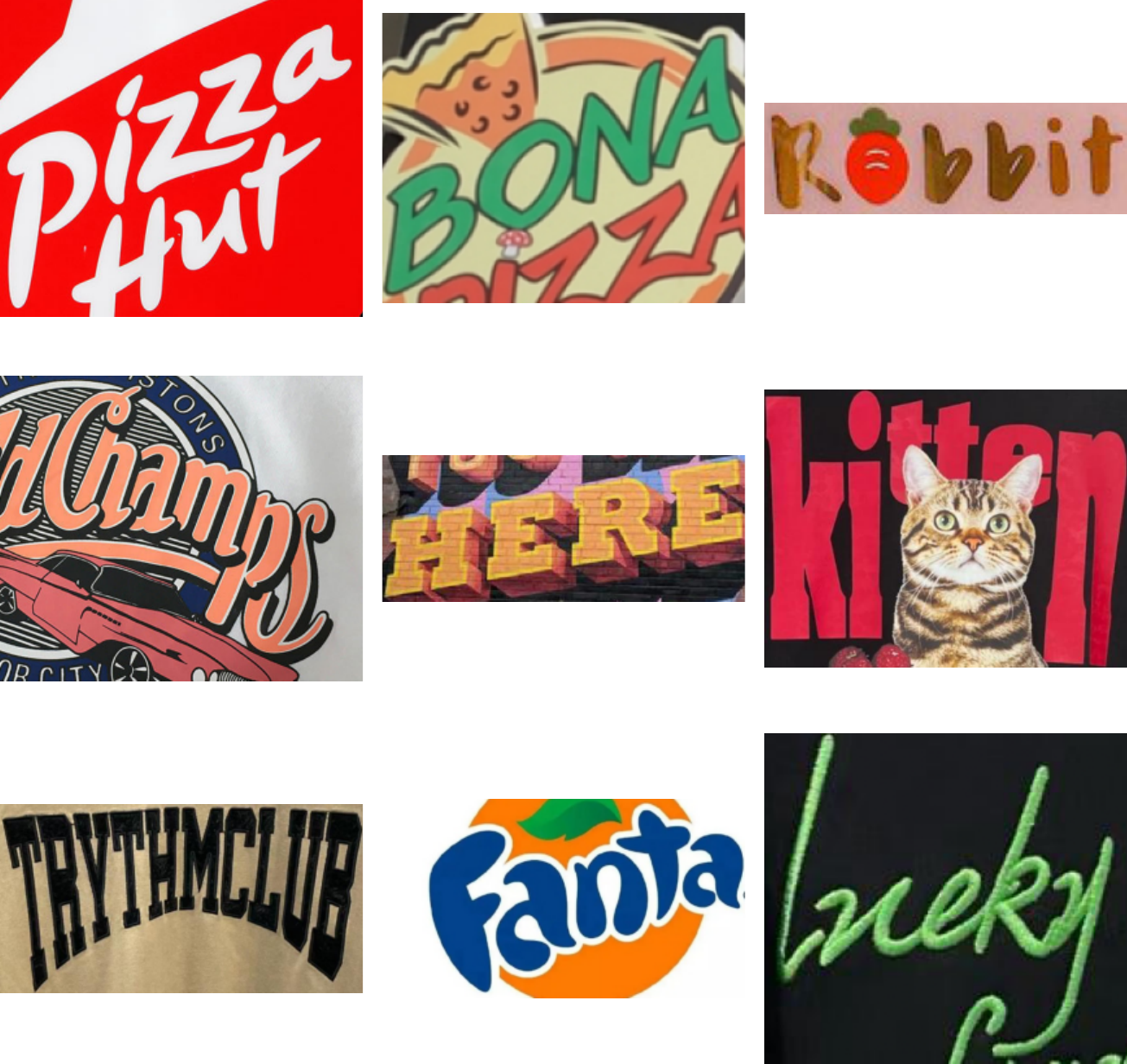}
    \vspace{-1.5mm}
    \caption{Examples from WordArt-V1.5, illustrating decorated characters, curved text, stylized fonts, and non-standard layouts that make \gls*{atr} substantially different from conventional scene text reading~\cite{xie2024}.}
    \label{fig:dataset-examples}
\end{figure}

\gls{atr} aims to read words whose appearance is deliberately shaped by visual design rather than by readability alone~\cite{xie2022,xie2024,do2025skeleton}.
Unlike regular scene text, artistic text may include non-standard fonts, strong color variation, objects replacing characters, and deformations that place samples close to the human readability limit, as shown in \cref{fig:dataset-examples}.
These properties make \gls*{atr} a stress test for modern recognizers because errors often arise from both visual ambiguity and language-level uncertainty.

The task is related to \gls{str}, but the domain gap is non-trivial.
Many \gls*{str} benchmarks focus on natural images affected by perspective distortion, blur, occlusion, or low resolution~\cite{wang2011,mishra2012,phan2013,karatzas2015,baek2019}, challenges also observed in specialized text-recognition applications such as license plate recognition~\cite{laroca2025advancing,laroca2026competition}.
WordArt-style images add difficulty because letters are embedded in artistic compositions, not merely degraded by acquisition conditions~\cite{xie2022,jiang2023}.
Thus, recognizers that perform well on regular benchmarks may still fail when characters are stretched, merged with drawings, stylized as objects, or intentionally distorted to create a visual identity.
This is relevant to posters, advertisements, logos, book covers, social media images, and other media where text and design are~inseparable.

The WordArt-V1.5 dataset, introduced in the \textit{ICDAR 2024 Competition on Artistic Text Recognition}~\cite{xie2024}, provides a standardized setting for this problem.
It contains 6,000 training images and a 6,000-image test set split evenly into Test~A and Test~B.
The fixed competition partitions enable controlled comparisons between general-purpose \gls*{str} recognizers, \gls*{atr}-oriented architectures, and ensembles, addressing the issue that inconsistent protocols can make model comparisons difficult to interpret~\cite{baek2019,xie2024}.
They also help separate model failures from annotation noise, visual ambiguity, and limited visual evidence, which is essential when some samples are difficult even for human~readers.

This paper studies WordArt-V1.5 as a reproducible benchmark for current recognizers and proposes a confidence-aware ensemble tailored to this setting.
Rather than introducing a new backbone, it addresses a practical question: \textit{how far can a compact fusion of strong recent recognizers go under fixed training, validation, and test partitions?}
We evaluate three widely adopted \gls*{str} models, MAERec~\cite{jiang2023}, SVTRv2~\cite{du2025svtrv2}, and PARSeq~\cite{bautista2022}, fine-tune them under the same protocol, combine them through a disagreement-aware confidence rule, and add a refinement stage for long words.
The long-word stage is motivated by the word-level metric, since longer words are more likely to contain at least one character error that invalidates the full~prediction.

The design separates three aspects of the problem.
First, the protocol evaluates representative recognizers under the same split, following the recommendation that \gls*{str} methods should be compared under consistent datasets, metrics, and implementation assumptions~\cite{baek2019}.
Second, the ensemble rule focuses on disagreement positions rather than averaging confidence across the whole word, prioritizing the characters that resolve competing predictions.
Third, the long-word module is activated only when word-level evaluation is more sensitive to insertions, deletions, and isolated low-confidence characters, reducing unnecessary dictionary intervention on short or confident~words.

The contributions are threefold.
First, we report a standardized benchmark of representative recognizers on WordArt-V1.5 using the official competition splits and \gls*{wra} metric.
Our code is available at \textit{\url{https://github.com/lucas-azdias/Artistic-Text-Recognition/}} to support reproducibility beyond the reported scores.
Second, we propose a confidence-aware ensemble with targeted long-word refinement based on sequence alignment and lexicon-guided correction.
Third, we provide an error analysis of all 303 remaining Test~B mistakes, separating model errors from labeling issues, ambiguity, and illegible samples.
Together, these results provide both a competitive recognition system and diagnostic evidence about the current limits of \gls*{atr} on~WordArt-V1.5.

The rest of this paper is organized as follows.
\cref{sec:related_work} reviews related work and the WordArt-V1.5 protocol.
\cref{sec:methodology} describes model selection, fine-tuning, the ensemble rule, and long-word refinement.
\cref{sec:results} reports quantitative results, ablations, and error analysis, while \cref{sec:conclusions} summarizes the findings, discusses limitations, and outlines future~directions.

%% file: 2-related.tex
\section{Related Work and WordArt-V1.5 Protocol}
\label{sec:related_work}

\glsreset{wra}

\subsection{Scene and Artistic Text Recognizers}

Scene text recognition has progressed from convolutional and recurrent sequence models toward transformer, masked-reconstruction, and refined \gls*{ctc}-based approaches~\cite{graves2006,shi2016crnn,baek2019}.
ASTER~\cite{shi2018aster} combined rectification and attention to handle irregular layouts, while ViTSTR~\cite{atienza2021} showed that a compact \gls*{vit} can recognize text without recurrent decoding or explicit rectification.

Recent methods are particularly relevant to \gls*{atr}, where stylization can substantially alter character shapes, spacing, and boundaries.
CornerTransformer~\cite{xie2022} employs contour-guided representations to recognize arbitrary and artistic text; PARSeq~\cite{bautista2022} explores multiple decoding orders through permutation language modeling; MAERec~\cite{jiang2023} learns robust visual representations through masked autoencoding; and SVTRv2~\cite{du2025svtrv2} improves \gls*{ctc}-based recognition through multi-size resizing, feature reorganization, and semantic guidance.
These complementary architectures provide a diverse set of candidates for both benchmarking and ensemble construction.
Their published results on widely used \gls*{str} benchmarks are summarized in \cref{tab:state-of-art-models}.

\begin{table}[!htb]
    \centering
    \caption{Representative recognizers considered in the benchmark.}
    \vspace{-1.5mm}
    \begin{threeparttable}
        \begin{tabular}{lccccc}
            \toprule
            \multirow{2}{*}{\textbf{Model}} & \multicolumn{5}{c}{\textbf{\acrfull*{wra} (\%)}} \\
            \cmidrule{2-6}
            & IIIT5k & SVT & IC15 & SVTP & U14M-B\textsuperscript{a} \\
            \midrule
            \acrshort{vitstr}~\cite{atienza2021} & 88.4 & 87.7 & 72.6 & 81.8 & -- \\
            CornerTransformer~\cite{xie2022} & 95.9 & 94.6 & 86.3 & 91.5 & -- \\
            \acrshort{parseq}~\cite{bautista2022} & 99.1 & 97.9 & 89.6 & 95.7 & \phantom{\textsuperscript{b}}84.3\textsuperscript{b} \\
            \acrshort{maerec}~\cite{jiang2023} & 98.5 & 97.8 & 89.5 & 94.4 & 85.2 \\
            \acrshort{svtrv2}~\cite{du2025svtrv2} & 99.2 & 98.0 & 91.1 & 99.0 & 86.1 \\
            \bottomrule
        \end{tabular}
        \begin{tablenotes}
            \footnotesize
            \item[a] Union14M-Benchmark.
            \item[b] Reported by Du \textit{et al.}~\cite{du2025svtrv2}, since Union14M-Benchmark was not available when PARSeq was published.
        \end{tablenotes}
    \end{threeparttable}
    \label{tab:state-of-art-models}
\end{table}

\subsection{WordArt-V1.5 Protocol}

WordArt-V1.5 extends the WordArt dataset introduced with CornerTransformer~\cite{xie2022} and was released for the \textit{ICDAR 2024 Competition on Artistic Text Recognition}~\cite{xie2024}.
It contains word images from posters, greeting cards, covers, billboards, advertisements, and other graphical designs, where artistic intent may alter character shape, spacing, orientation, and texture.
The adopted partitions are summarized in \cref{tab:protocol}.

\begin{table}[!htb]
    \centering
    \caption{WordArt-V1.5 protocol used in this paper.}
    \vspace{-1.5mm}
    \begin{tabular}{lcc}
        \toprule
        \textbf{Split} & \textbf{Images} & \textbf{Use in this paper} \\
        \midrule
        Train & 6,000 & Fine-tuning \\
        Test~A & 3,000 & Validation and model selection \\
        Test~B & 3,000 & Final evaluation \\
        \bottomrule
    \end{tabular}
    \label{tab:protocol}
\end{table}

\begin{figure}[!htb]
    \centering
    \begin{minipage}[t]{0.48\linewidth}
        \centering
        \includegraphics[width=\linewidth]{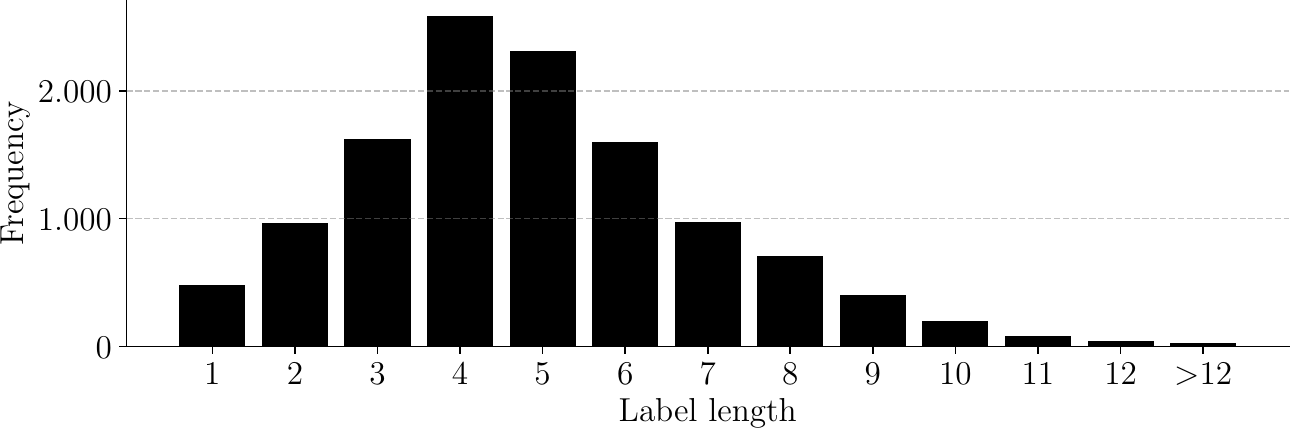}
        \small (a) Label length distribution.
    \end{minipage}
    \hfill
    \begin{minipage}[t]{0.48\linewidth}
        \centering
        \includegraphics[width=\linewidth]{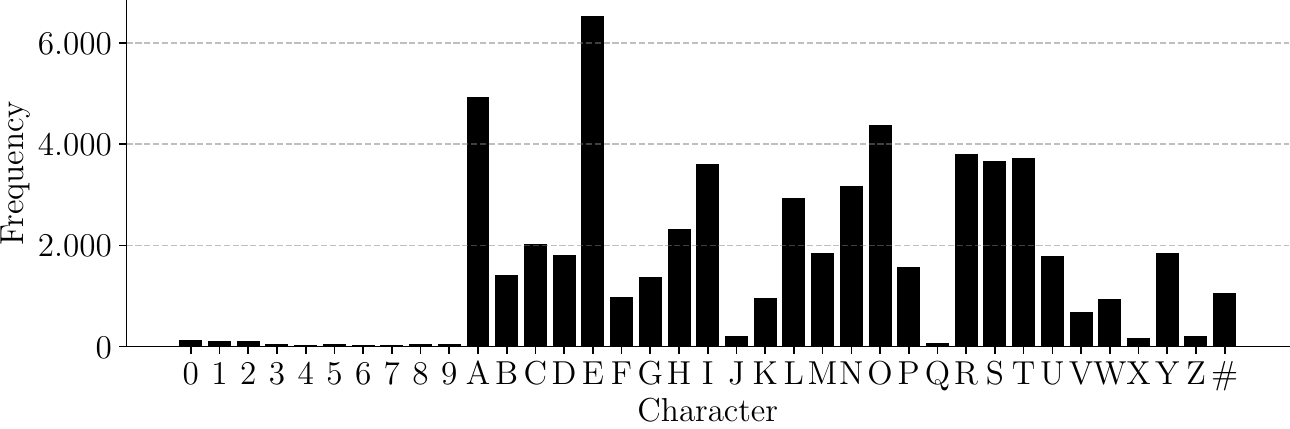}
        \small (b) Character frequency distribution.
    \end{minipage}

    \caption{WordArt-V1.5 label statistics, with symbols grouped into the \texttt{\#} class.}
    \label{fig:dataset-statistics}
\end{figure}

The official metric is \gls{wra}, which ignores case and symbols and counts a prediction as correct only when the normalized full word matches the ground truth~\cite{xie2024}.
Let $W$ denote the number of evaluated words and $W_r$ the number of correctly recognized words:
\begin{equation}
    WRA=\frac{W_r}{W} \, .
    \label{eq:wra}
\end{equation}
Any incorrect, missing, or inserted character invalidates the full prediction. Hence, longer labels provide more opportunities for word-level errors, motivating the analysis in \cref{fig:dataset-statistics}.

We preserve the official partitions: public pretrained checkpoints are fine-tuned on Train, Test~A is used for validation and model selection, and Test~B is reserved for final reporting.
The public leaderboard is used only for context, not as evidence of an official competition entry.

\subsection{Public Baselines on WordArt-V1.5}

Public WordArt-V1.5 systems frequently combine complementary models and objectives~\cite{xie2024}.
Examples include supervised, semi-supervised, and self-supervised components in \textit{OCR for WordArt}; ensembling and character-to-character distillation in \textit{ViettelAI-OCR}; a character-level contrastive adaptation of MAERec in \textit{Let Me See}; and a PARSeq, MAERec, and CornerTransformer ensemble in \textit{iPad\_OCR}.
Rather than reproducing a large leaderboard-oriented pipeline, we construct a compact and reproducible baseline using public checkpoints, a fixed fine-tuning protocol, and a transparent fusion~rule.

%% file: 3-method.tex
\section{Confidence-Aware Ensemble}
\label{sec:methodology}

\subsection{Candidate Models and Fine-Tuning}

\begin{figure}[!htb]
    \centering
    \includegraphics[width=0.70\linewidth]{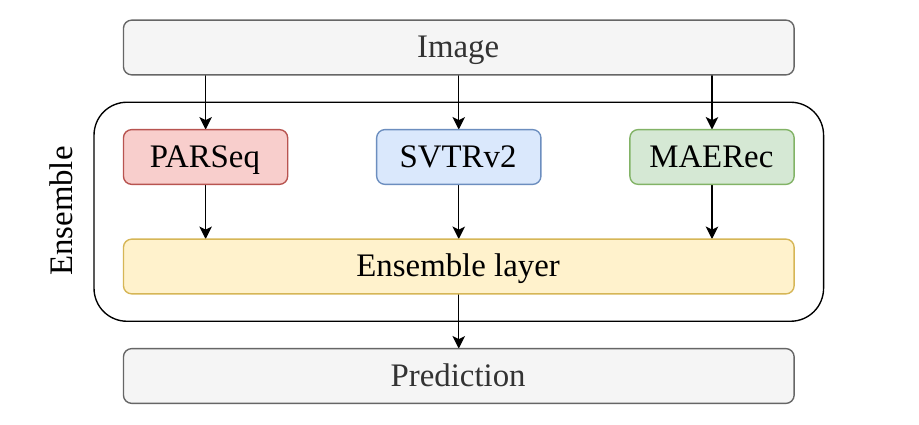}
    \vspace{-1.5mm}
    \caption{Overview of the proposed ensemble, which combines PARSeq, SVTRv2, and MAERec through a confidence-aware aggregation layer.}
    \label{fig:solution}
\end{figure}

The proposed system, shown in \cref{fig:solution}, combines three recognizers selected from five candidates.
All candidates were first evaluated on Test~A using pretrained checkpoints, as reported in \cref{tab:model-selection}.
The strongest and most complementary models were SVTRv2~\cite{du2025svtrv2}, PARSeq~\cite{bautista2022}, and MAERec~\cite{jiang2023}, with Test~A \gls*{wra} values of 85.80\%, 83.97\%, and 83.40\%, respectively.
ViTSTR~\cite{atienza2021} and CornerTransformer~\cite{xie2022} reached 78.80\% and 73.80\%, and were not included in the final ensemble.
This keeps the ensemble compact while preserving diversity across \gls*{ctc}, autoregressive, and masked-reconstruction-based recognizers.

\begin{table}[!htb]
    \centering
    \caption{Candidate model performance on Test~A, used as the validation set in this paper, before and after fine-tuning.}
    \vspace{-1.5mm}
    \resizebox{0.99\linewidth}{!}{
    \begin{tabular}{lcc}
        \toprule
        \textbf{Model} & \textbf{Pretrained WRA (\%)} & \textbf{Fine-tuned WRA (\%)} \\
        \midrule
        SVTRv2~\cite{du2025svtrv2} & 85.80 & 86.23 \\
        PARSeq~\cite{bautista2022} & 83.97 & 84.53 \\
        MAERec~\cite{jiang2023} & 83.40 & 84.53 \\
        ViTSTR~\cite{atienza2021} & 78.80 & -- \\
        CornerTransformer~\cite{xie2022} & 73.80 & -- \\
        \bottomrule
    \end{tabular}}
    \label{tab:model-selection}
\end{table}

\begin{figure}[!htb]
    \centering
    \includegraphics[width=0.50\linewidth]{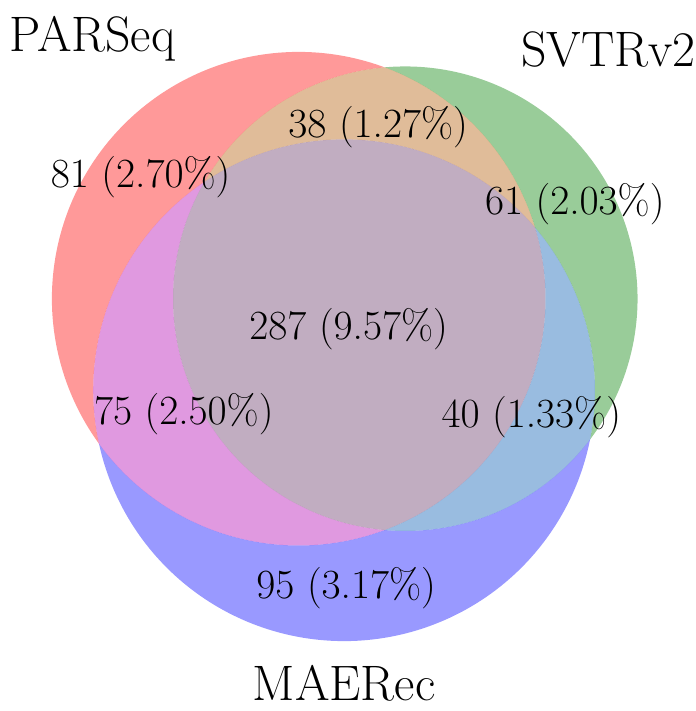}
    \vspace{-1.5mm}
    \caption{Error overlap among the selected pretrained models on Test~A.}
    \label{fig:venn-pretrained}
\end{figure}

The decision to use an ensemble is supported by the Test~A error overlap in \cref{fig:venn-pretrained}.
Among 677 samples missed by at least one selected model, only 287 errors were shared by all three models, corresponding to 42.39\% shared errors.
If an oracle could always choose a correct prediction whenever at least one model was right, the theoretical Test~A upper bound would be 90.43\%, substantially higher than any individual model.
This gap indicates that many mistakes are model-specific rather than inherent to all recognizers.
It motivated a fusion rule that focuses on disagreement positions rather than average confidence over the full sequence.

\major{Each selected recognizer was fine-tuned on the official training split, while Test~A was used exclusively for validation, hyperparameter selection, and checkpoint selection. The main configurations selected through this validation procedure are summarized in \cref{tab:finetuning-config}. All models were optimized with AdamW and trained for at most 50 epochs, with the best checkpoint selected according to the WRA obtained on Test~A.}

\begin{table}[!htb]
    \centering
    \caption{\major{Main fine-tuning configurations selected based on validation performance on Test~A. All models used the AdamW optimizer.}}
    \label{tab:finetuning-config}
    \vspace{-1.5mm}
    \resizebox{0.95\linewidth}{!}{
    \begin{tabular}{lccc}
        \toprule
        \textbf{Configuration} & \textbf{MAERec} & \textbf{PARSeq} & \textbf{SVTRv2} \\
        \midrule
        Initial learning rate
            & $1\times10^{-5}$
            & $7\times10^{-4}$
            & $5\times10^{-5}$ \\
        Batch size
            & 64
            & 384
            & 128 \\
        Scheduler
            & CosineAnnealingLR
            & OneCycleLR
            & CosineAnnealingLR \\
        \bottomrule
    \end{tabular}}
\end{table}

\major{Fine-tuning employed online geometric and photometric augmentations to reduce sensitivity to distortions and appearance variations~\cite{shi2018aster,baek2019}. These augmentations included rotations, translations, scaling, shearing, affine and perspective transformations, pyramid resizing, photometric and color perturbations, Gaussian noise, motion blur, generic blur, and random erasing. After fine-tuning, SVTRv2 improved from 85.80\% to 86.23\% WRA on Test~A, while PARSeq and MAERec both reached 84.53\%. These validation results guided the selection of the final models and configurations, without using Test~B for parameter tuning.}

Experiments were implemented in Python with PyTorch and executed on Google Colab with an NVIDIA L4 GPU, 22.5 GB of memory, and 53 GB of system memory.
This environment was sufficient to fine-tune the selected recognizers and evaluate the ensemble without multi-GPU infrastructure.

\subsection{Disagreement-Aware Confidence Rule}

Let $M=\{m_1,m_2,\ldots,m_K\}$ be the set of recognizers.
Each recognizer $m_k$ produces a predicted sequence $\hat{\mathbf{y}}_k=[\hat{y}_{k,1},\ldots,\hat{y}_{k,n}]$ and confidence scores $\mathbf{l}_k=[l_{k,1},\ldots,l_{k,n}]$.
We first identify positions at which at least two models disagree:
\begin{equation}
    D = \{j \mid \exists\,p,q,\; \hat{y}_{p,j} \neq \hat{y}_{q,j}\} \; .
    \label{eq:disagreement}
\end{equation}

The score of model $m_k$ is the minimum confidence over disagreement positions.
When all predictions agree, the minimum over the whole sequence is used:
\begin{equation}
    S_k =
    \begin{cases}
        \min_{j \in D} l_{k,j}, & \text{if } |D| > 0, \\
        \min_j l_{k,j}, & \text{otherwise.}
    \end{cases}
    \label{eq:min-score}
\end{equation}

The ensemble output is the sequence predicted by the model with the largest score:
\begin{equation}
    \hat{\mathbf{y}}_{ens}=\hat{\mathbf{y}}_{k^*},\quad k^*=\arg\max_k S_k \; .
    \label{eq:ensemble}
\end{equation}

This rule penalizes low-confidence characters exactly where recognizers disagree, which is more targeted than selecting a prediction by global average confidence.
A model can therefore keep a high score when it is uncertain only on characters shared by all predictions, but it is penalized when uncertainty occurs at a position that changes the final word.
Scores are normalized by each model's mean character confidence to reduce calibration differences across recognizers, since modern neural networks can produce confidence values that are not directly comparable across architectures~\cite{guo2017,laroca2023leveraging}.
The rule remains conservative: by default, it selects one complete model hypothesis rather than synthesizing a new word through character-level voting.
This avoids creating implausible sequences from locally confident but globally inconsistent characters.
\major{To assess the contribution of this scoring rule, \cref{tab:strategy-ablation} compares the proposed minimum-weighted strategy with majority voting and maximum-weighted selection on Test~A.
The minimum-weighted strategy achieves the highest WRA, supporting the use of the least-confident disagreement position to select among the complete-model~hypotheses}.

\begin{table}[!htb]
    \centering
    \caption{Ablation of ensemble scoring strategies on Test~A.}
    \vspace{-1.5mm}
    \begin{tabular}{lcc}
        \toprule
        \textbf{Strategy} & \textbf{WRA (\%)} \\
        \midrule
        Majority voting & 85.67 \\
        Maximum weighted & 87.20 \\
        \textbf{Minimum weighted} & \textbf{87.77} \\
        \bottomrule
    \end{tabular}
    \label{tab:strategy-ablation}
\end{table}

\subsection{Long-Word Refinement}

The full-word metric makes long labels sensitive to isolated character errors.
If each character has an independent error probability $p$, the probability of a correct word of length~$n$ is~$(1-p)^n$.
Thus, small character-level uncertainty accumulates quickly as $n$ increases.
In WordArt-V1.5, labels with at least nine characters represent 749 samples in the full dataset and 184 samples in Test~B.

For words with at least nine characters, we apply a two-stage refinement using a threshold chosen empirically on the validation set.
First, the predictions of SVTRv2, PARSeq, and MAERec are aligned with the Needleman-Wunsch algorithm~\cite{willis2009,eger2013}.
The aligned sequences enable character-level voting even when insertions or deletions shift positions across models.
Gap positions inherit the confidence of the nearest available neighboring character.
Second, when the aligned word contains at least one character with confidence below 60\%, a lexicon correction is considered.
The lexicon stage searches the \texttt{wlist\_match3} English word list~\cite{vertanen} using bigram-based Levenshtein similarity~\cite{kondrak2005}.
A correction is accepted only when the best candidate reaches at least 70\% similarity, reducing the risk of replacing a confident artistic word with an unrelated dictionary entry, a known concern in unconstrained scene text~\cite{baek2019}.

The refinement is not post-processing applied to every output.
It targets the subset where the metric, the label-length distribution, and the observed disagreements indicate a specific weakness.
This design is important because dictionary correction can be harmful when the image contains a brand name, a stylized invented word, or a proper noun that is absent from the lexicon.
Restricting the module to long words with low-confidence evidence helps recover likely character-level mistakes while preserving the original model prediction in most~cases.

%% file: 4-results.tex
\section{Results and Analysis}
\label{sec:results}

\subsection{Overall Recognition Accuracy}

\cref{tab:testb-results} reports the final Test~B results.
The proposed ensemble reaches 89.90\% \gls*{wra}, outperforming the best individual fine-tuned model by 1.77 percentage points.
SVTRv2 remains the strongest individual model, but PARSeq and MAERec provide alternative hypotheses that the ensemble can exploit when SVTRv2 is uncertain.
The contextual comparison with public WordArt-V1.5 systems shows that the compact ensemble is within the range of strong specialized pipelines.

\begin{table}[!htb]
    \centering
    \caption{Final Test~B results and contextual WordArt-V1.5 baselines.}
    \vspace{-1.5mm}
    \resizebox{0.99\linewidth}{!}{
    \resizebox{!}{7.5ex}{
    \begin{tabular}{lc}
        \toprule
        \textbf{Method} & \textbf{\gls*{wra} (\%)} \\
        \midrule
        MAERec~\cite{jiang2023} fine-tuned & 87.50 \\
        PARSeq~\cite{bautista2022} fine-tuned & 87.60 \\
        SVTRv2~\cite{du2025svtrv2} fine-tuned & 88.13 \\
        \textbf{Proposed ensemble} & \textbf{89.90} \\
        \bottomrule
    \end{tabular}
    }
    \resizebox{!}{7.5ex}{
    \begin{tabular}{lc}
        \toprule
        \textbf{Method} & \textbf{\gls*{wra} (\%)} \\
        \midrule
        iPad\_OCR~\cite{xie2024} & 89.27 \\
        Let Me See~\cite{xie2024} & 89.77 \\
        ViettelAI-OCR~\cite{xie2024} & 90.77 \\
        Ocr For WordArt~\cite{xie2024} & 91.07 \\
        \bottomrule
    \end{tabular}
    }}
    \label{tab:testb-results}
\end{table}

This comparison is informative because all entries use the same Test~B split and \gls*{wra} metric, the type of controlled protocol needed for meaningful text-recognition comparisons~\cite{baek2019,xie2024}.
The proposed system is not intended to replace larger leaderboard-oriented pipelines.
Instead, it provides a transparent baseline for future \gls*{atr} studies and shows that a compact three-model system can recover a relevant portion of the errors left by a single strong recognizer.
In this sense, the main result is not only the final score, but also the gap between individual fine-tuned models and the confidence-aware ensemble under the same validation and test protocol.

\begin{figure}[!htb]
    \centering
    \resizebox{\linewidth}{!}{
    \begin{minipage}[t]{0.5\linewidth}
        \centering
        \includegraphics[width=\linewidth]{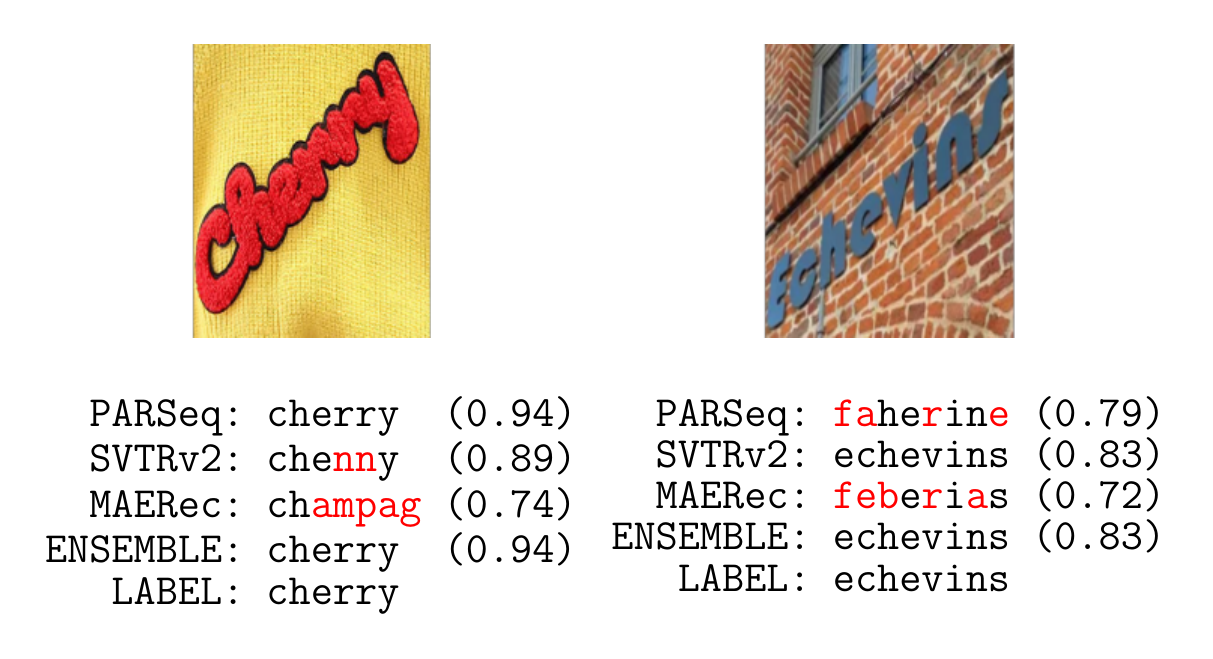}
        \scriptsize (a) Correct predictions.
    \end{minipage}\hspace{-2mm}
    \begin{minipage}[t]{0.5\linewidth}
        \centering
        \includegraphics[width=\linewidth]{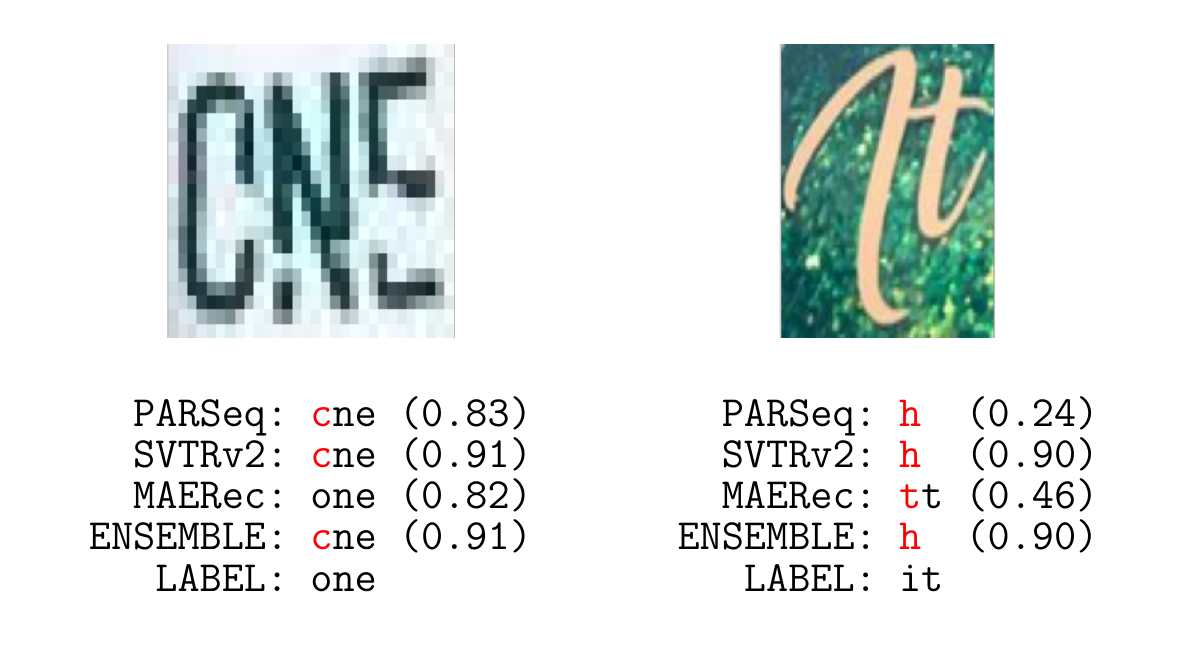}
        \scriptsize (b) Incorrect predictions.
    \end{minipage}}

    \caption{Qualitative examples from the final ensemble, showing
(a)~cases where confidence-aware fusion selects the correct prediction and
(b)~cases where all models remain confused.}
    \label{fig:qualitative-results}
\end{figure}

\cref{fig:qualitative-results} illustrates both successful and failed cases.
In some samples, the ensemble selects a correct hypothesis even when another recognizer produces a visually similar but wrong word.
In others, all recognizers converge to the same wrong word or assign high confidence to an incorrect interpretation, showing that many remaining errors arise from stylized shapes with multiple plausible readings.

\subsection{Effect of Long-Word Refinement}

The refinement stage is intentionally targeted, so its global gain is modest.
As shown in \cref{tab:ablation}, the complete refinement improves Test~B \gls*{wra} from 89.67\% to 89.90\%, a gain of 0.23 percentage points over all 3,000 test samples.
Within the 184-sample long-word subset, however, the gain is 2.72 percentage points.
This targeted improvement is meaningful because long artistic words concentrate a difficult failure mode: a visually plausible prediction can be rejected by \gls*{wra} due to a single shifted, inserted, or low-confidence character.

\begin{table}[!htb]
    \centering
    \caption{Ablation of the long-word refinement on Test~B.}
    \vspace{-1.5mm}
    \resizebox{\linewidth}{!}{
    \begin{threeparttable}
        \begin{tabular}{lcc}
            \toprule
            \textbf{Refinement} & \textbf{WRA (\%)} & \textbf{Long-word WRA\textsuperscript{a} (\%)} \\
            \midrule
            None & 89.67 & 75.00 (--)\phantom{abdefIii} \\
            Needleman-Wunsch only & 89.73 & 75.00 (+0.00 pp) \\
            Dictionary only & 89.80 & 76.63 (+1.63 pp) \\
            Needleman-Wunsch + dictionary & \textbf{89.90} & \textbf{77.72 (+2.72 pp)} \\
            \bottomrule
        \end{tabular}
        \begin{tablenotes}
            \footnotesize
            \item[a] Gain measured only on Test~B words with at least nine characters.
        \end{tablenotes}
    \end{threeparttable}
    }
    \label{tab:ablation}
\end{table}

\major{The minimum word-length threshold was selected through an ablation on Test~A. Increasing the threshold from three to eight characters progressively reduced the refinement coverage from 91.13\% to 11.50\%, while increasing the WRA from 87.00\% to 87.93\%. We selected a threshold of nine characters, corresponding to 5.70\% coverage and 87.73\% WRA, to restrict the module to the intended long-word cases. Higher thresholds of 10 and 11 characters reduced coverage to 2.53\% and 1.13\%, respectively, without providing consistent accuracy gains.}

The ablation should therefore be interpreted on its target subset.
As the module is activated for only 6.13\% of Test~B, even a substantial correction rate inside that subset appears small when averaged across all samples.
This is expected and should not be interpreted as evidence that the module is ineffective.
Needleman-Wunsch alignment is most helpful when models disagree on insertions or deletions, while the dictionary stage is most useful when the aligned result contains a localized low-confidence character.
The combined alignment and dictionary strategy improves the intended cases without degrading the remaining~predictions.

\subsection{Qualitative and Error Analysis}

\begin{figure}[!htb]
    \centering
    \includegraphics[width=0.80\linewidth]{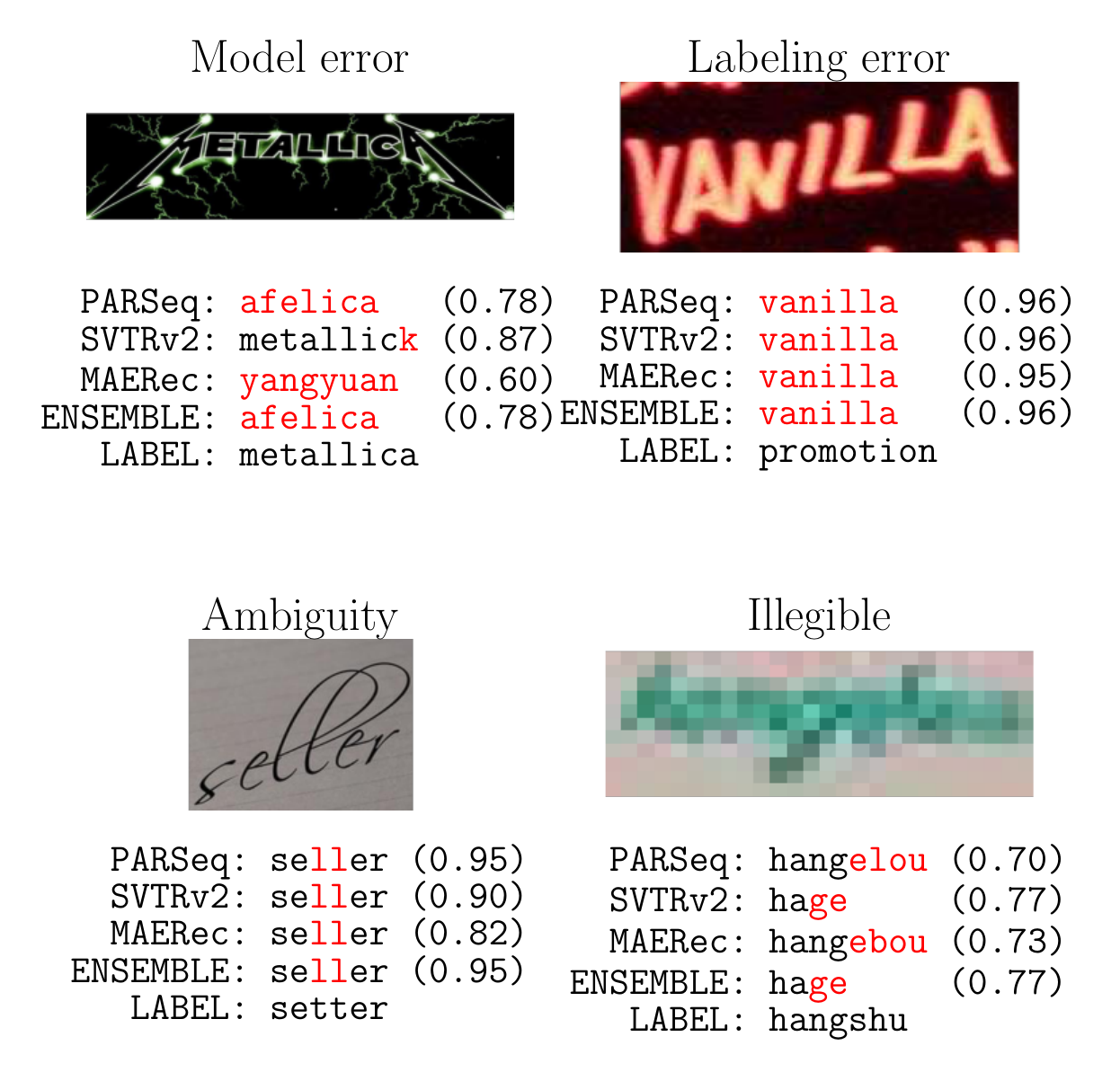}
    \vspace{-3mm}
    \caption{Examples from the manual error analysis, covering model errors, labeling errors, ambiguous words, and visually illegible samples.}
    
    \label{fig:results-problem}
\end{figure}

\cref{fig:results-problem} shows representative error cases, and \cref{tab:error-analysis} summarizes all 303 Test~B mistakes.
Model errors account for 51.2\% and mostly involve cursive characters, heavy stylization, missing strokes, or misleading decorative components.
Interestingly, the remaining 48.8\% are not purely model failures: \textit{labeling errors} account for 33.0\%, ambiguity for 10.9\%, and illegibility for 5.0\%.
This result shows why diagnostic reporting matters, since the same \gls*{wra} score may combine solvable recognition failures with samples whose ground truth or visual evidence is debatable.

\begin{table}[!htb]
    \centering
    \caption{Manual classification of the 303 Test~B errors of the proposed ensemble.}
    \vspace{-1.5mm}
    \begin{tabular}{lcc}
        \toprule
        \textbf{Error type} & \textbf{Count} & \textbf{Share (\%)} \\
        \midrule
        Model error & 155 & 51.2 \\
        Labeling error & 100 & 33.0 \\
        Ambiguity & \phantom{0}33 & 10.9 \\
        Illegible sample & \phantom{0}15 & \phantom{0}5.0 \\
        \bottomrule
    \end{tabular}
    \label{tab:error-analysis}
\end{table}

The labeling-error group is important for benchmark interpretation.
When the visual content clearly differs from the label, a correct model prediction may still be counted as an error by the official metric.
Ambiguous samples create a different problem, where the annotation may be plausible but another reading is also visually defensible.
Illegible samples provide insufficient evidence for a reliable word-level decision.
Separating these categories does not invalidate WordArt-V1.5; it clarifies its limits and helps prevent attributing every remaining error to model~weaknesses.

The analysis suggests that WordArt-V1.5 is valuable not only for ranking methods but also as a diagnostic benchmark.
It exposes complementary model errors, reveals the fragility of word-level scoring on long labels, and contains ambiguous cases that motivate uncertainty-aware evaluation.
Future reporting should therefore include official \gls*{wra}, targeted subset ablations, and qualitative error categories whenever possible.
Such reporting makes progress easier to interpret, especially when new methods obtain small global gains on a dataset where a non-negligible fraction of errors may be annotation-related or visually underdetermined.

%% file: 6-conclusions.tex
\section{Conclusions}
\label{sec:conclusions}

This paper presented a confidence-aware ensemble for Artistic Text Recognition (ATR) on WordArt-V1.5.
The system combines SVTRv2, PARSeq, and MAERec after fine-tuning, selects predictions through a disagreement-aware confidence rule, and applies targeted refinement to long words using sequence alignment and lexicon-guided correction.
On Test~B, the proposed ensemble achieved 89.90\% WRA, improving the best individual fine-tuned model by 1.77 percentage points.
Although the long-word refinement yielded a modest global improvement, it increased accuracy by 2.72 percentage points on its targeted subset, showing the benefit of addressing isolated character errors under word-level evaluation.

The error analysis also highlights limitations that are not captured by recognition accuracy alone.
Among the 303 remaining Test~B errors, 48.8\% were associated with labeling issues, visual ambiguity, or illegible samples.
These findings suggest that progress in ATR depends not only on stronger recognizers, but also on better confidence calibration, uncertainty estimation, language-aware decoding, and diagnostic evaluation, particularly for samples near the limits of human readability~\cite{baek2019,guo2017}.

Future work will explore learned fusion strategies and the use of generative models to expand ATR training data with controlled styles, long words, rare characters, and challenging distortions~\cite{gupta2016,rombach2022,chen2024textdiffuser2}.
Vision-Language Models (VLMs) may further support this process by filtering generated samples according to readability and label consistency~\cite{radford2021,liu2023llava}.
Future benchmarks could also report complementary measures such as long-word accuracy and uncertainty-aware rejection, helping distinguish genuine recognition failures from ambiguous or low-readability cases.

%% file: 0-acknowledgments.tex
\section*{\uppercase{Acknowledgments}}

\iffinal
    The authors thank the \textit{Pontifícia Universidade Católica do Paraná}~(PUCPR) for the financial support that made their participation in the conference possible.
\else
    \noindent\textit{The acknowledgments are hidden for review. The space below is reserved for the acknowledgments in the final version.}
    \vspace{2\baselineskip}
\fi